\documentclass{article}
\usepackage{ijcai16}
\usepackage{hyperref}
\usepackage{graphicx}
\usepackage{float}
\usepackage{caption}
\usepackage{subcaption}
\usepackage{lipsum}
\usepackage{amsfonts}
\hypersetup{
    colorlinks=true,
    urlcolor=black,
    citecolor=black,
}

\newcommand\blfootnote[1]{%
  \begingroup
  \renewcommand\thefootnote{}\footnote{#1}%
  \addtocounter{footnote}{-1}%
  \endgroup
}

\usepackage{times}

\title{Dynamic Frame skip Deep Q Network}
\author{Aravind S. Lakshminarayanan$^{*}$, Sahil Sharma$^{*}$, Balaraman Ravindran \\ 
RISE IIL, Indian Institute of Technology, Madras  \\
aravindsrinivas@gmail.com, ssahil08@gmail.com, ravi@cse.iitm.ac.in}

\begin{document}

\maketitle
\blfootnote{$^{*}$ Equal Contribution}

\begin{abstract}

  Deep Reinforcement Learning methods have achieved state of the art performance in learning control policies for the games in the Atari $2600$ domain. One of the important parameters in the Arcade Learning Environment (ALE, \cite{ale:scheme}) is the frame skip rate. It decides the granularity at which agents can control game play. A frame skip value of $k$ allows the agent to repeat a selected action $k$ number of times. The current state of the art architectures like Deep Q-Network (DQN, \cite{dqn:scheme}) and Dueling Network Architectures (DuDQN, \cite{dudqn:scheme}) consist of a framework with a static frame skip rate, where the action output from the network is repeated for a fixed number of frames regardless of the current state. In this paper, we propose a new architecture, Dynamic Frame skip Deep Q-Network (DFDQN) which makes the frame skip rate a dynamic learnable parameter. This allows us to choose the number of times an action is to be repeated based on the current state. We show empirically that such a setting improves the performance on relatively harder games like Seaquest.
  
\end{abstract}

\section{Introduction}
  
  Video game domains such as Mario and Atari $2600$ have served as a test bed to measure performance of learning algorithms in Artificial Intelligence (\cite{ale:scheme}). Such games present a unique challenge to an AI agent since performing well in them requires being able to understand the current state of the game (involves pattern recognition) as well as being able to choose actions considering long term reward (involves planning). Recent application of Deep Learning to the Atari $2600$ domain has resulted in state of the art performance. The Deep Q-learning neural network architecture \cite{dqn:scheme} is one such method that uses a deep neural network with convolutional filters to learn a hierarchically compositional representation of the state space. Such a representation is used to approximate the state-action value function and more importantly generalize to unseen states.

  One of the hyper-parameters used in the DQN framework is the {\it frame skip rate} which denotes the number of times an action selected by the agent is to be repeated. If the frame skip rate is low, the decision making frequency of the agent is high. This leads to policies which change rapidly in space and time. It also leads to slower game play since the DQN has to make a decision more often and each such decision involves running convolutional neural networks which are computationally intensive. On the other hand, a high frame skip rate causes infrequent decisions, which reduces the time to train on an episode at the cost of losing fine-grained control. 
  The policies learnt with a higher frame skip rate have fewer action sequences which exhibit super-human reflexes as compared to a DQN with a low frame skip rate (\cite{dqn:scheme} use a frame skip rate of $4$). A reduction in decision making frequency gives the agent the advantage of not having to learn control policies in the intermediate states given that a persistent action from the current state can lead the agent to an advantageous next state. Such skills would be useful for the agent in games where good policies require some level of temporal abstraction. An example of this situation in Seaquest is when the agent has to continually shoot at multiple enemies arriving together and at the same depth, one behind another. 
  Having said that, there are also situations where the agent has to take actions which require quick reflexes to perform well with an example from Seaquest being states in which multiple enemies attack at once and are placed at varying depths. The agent is then expected to manoeuvre skillfully to avoid the enemies.
  Therefore, a high but static frame skip rate is probably not a solution for better game playing strategies in spite of the temporal abstraction it provides. 
  
  As a first step in the direction of temporal abstractions in the policy space, we explore a dynamic frame skip model, built on top of the DQN architecture. In this model, the agent can choose to play from a set of options or macro actions corresponding to two different (and fixed) frame skip rate values available at every state of the game. We show the efficacy of the policies learnt using such a model over those learnt using the DQN model (which has a static frame skip) by empirically establishing the fact that dynamic action repetition is an important way in which an AI agent's capabilities can be extended. Thereby, we make the claim that there is a need to explore and experiment with structured parametrized policies using Actor Critic setup in the Atari $2600$ domain. In the model that we propose, the agent decides whether to take a repetitive action sequence that is temporally long (corresponding to a higher frame skip) or a shorter sequence of repeated actions (corresponding to a lower frame skip) based on the input image. We also outline a generic method wherein the agent's policy for a particular frame would not only contain the action probabilities, but the level of persistence (frame skip) of the actions as well. Note that no experiments are performed for the generic method and only the framework is proposed. 

\section{Related work}

One of the first efforts which pushed the state of the art significantly in the Atari $2600$ domain was \cite{dqn:scheme}. Their architecture (DQN) motivated the use of convolutional neural networks for playing games in the Atari $2600$ domain. DQN combines the 
image-based-state representational capabilities of deep convolutional neural networks with the robustness of a Q-learning set up. Our work is a direct modification of the DQN architecture.

\cite{fs:scheme} is one work that focuses on the power of the frame skip rate parameter with experiments in the Atari $2600$ domain. Their learning framework is a variant of Enforced Sub-populations (ESP), a neuroevolution approach which has been successfully trained for complex control tasks such as controlling robots and playing games. They show empirically that the frame skip parameter is an important factor deciding the performance of their agent in the Atari domain. They demonstrate for example that Seaquest achieves best performance when a frame skip rate of $180$ frames is used. This is equivalent to the agent pausing (continuing the same action) for three seconds between decisions since the ALE emulator runs at $60$ frames per second.  
They argue that a higher value of frame skip rate allows the agent to not learn action selection for the states that are skipped and hence develop associations between states and actions that are temporally distant. However, they experiment only with a static frame skip setup. The key way in which DFDQN differs from both the ESP and the DQN approaches is that we make frame skip a dynamic learnable parameter.

The idea of dynamic length temporal abstractions in the policy space on Atari Domain has been explored by \cite{vafa:scheme}. They use a Monte Carlo Tree Search (MCTS) planner with macro-actions that are composed of the same action repeated $k$ times, for different $k$. 
The way in which our approach differs from \cite{vafa:scheme} is in terms of using DQN to build neural network Q-value approximators instead of making use of search techniques that cannot generalize. This leads to not only a high scoring AI agent but also one which is capable of playing at human speeds because of generalization to states not encountered before.

Similar to us, \cite{Ortega:scheme} attempt to design high level actions to imitate human-like play in Mario. They compose some high level actions based on human play like {\it Walk, Run, Right Small Jump, Right High Jump} and {\it Avoid enemy}, which are macro actions composed of varying length repetitive key presses. For instance, {\it Right High Jump} involves pressing the {\it Jump and Right} keys together for $10$ frames, while {\it Right Small Jump} requires the same for $2$ frames.




\section {Background}
\subsection{Q Learning}
Control policies for game environments have commonly been learnt through the Reinforcement Learning framework \cite{silver:scheme,dqn:scheme,Narasimhan:scheme}. The game is formulated as a sequence of transition tuples $(s,a,r,s')$ in a Markov Decision Process (MDP). The state-action value $Q$ function is used to guide the agent to selecting the optimal action $a$ at a state $s$, where $Q(s,a)$ is a measure of the long-term reward obtained by taking action $a$ at state $s$. One way to learn optimal policies for an agent is to estimate the optimal $Q(s,a)$ for the task. Q-learning (\cite{Watkins:scheme}) is an off-policy Temporal Difference (TD) learning (\cite{suttonbarto:scheme}) algorithm that does so. The Q-values are updated iteratively through the Bellman optimality equation (\cite{suttonbarto:scheme}) with the rewards obtained from the game as below:
$$Q_{i+1}(s,a) = \mathbb{E}[r + \gamma \textrm{argmax}_{a'} Q_i (s',a') | (s,a)]$$ 
During the update process, the agent chooses the action with the highest $Q(s,a)$ at each state $s$ while playing the game, though, for the sake of exploration, the agent executes an $\epsilon$ greedy behavioural policy.

\subsection{Deep Q Network (DQN)}
In high dimensional state spaces, it is infeasible to update Q-value for all possible state-action pairs. One way to address this issue is by approximating $Q(s,a)$ through a parametrized function approximator $Q(s,a;\theta)$, thereby generalizing over states and actions by operating on higher level features (\cite{suttonbarto:scheme}). To be able to discover these features without feature engineering, we need a neural network function approximator. The DQN (\cite{dqn:scheme}) approximates the Q-value function with a deep neural network to be able to predict $Q(s,a)$ over all actions $a$, for all states $s$. 
The loss function used for learning a Deep Q Network is as below:
$$L_i(\theta_i) = \mathbb{E}_{s,a,r,s'}[(y_i^{DQN}- Q(s,a;\theta_i))^2],$$
with
$$y_i^{DQN} = (r_i+ \gamma \textrm{max}_{a'}Q(s',a',\theta^{-}))$$
Here, $L_i$ represents the expected TD error corresponding to current parameter estimate $\theta_i$. $\theta^{-}$ represents the parameters of a separate {\it target network}, while $\theta_i$ represents the parameters of the {\it online network}. The usage of a {\it target network} is to improve the stability of the learning updates. The gradient descent step is shown below:
$$\nabla_{\theta_i} L_i(\theta_i) = \mathbb{E}_{s,a,r,s'} [(y_i^{DQN} - Q(s,a;\theta_i))\nabla_{\theta_i} Q(s,a)]$$

To avoid correlated updates from learning on the same transitions that the current network simulates, an experience replay (\cite{Lin:scheme}) $D$ (of fixed maximum capacity) is used, where the experiences are pooled in a FIFO fashion. Transitions from previous episodes are sampled multiple times to update the current network, thereby avoiding divergence issues due to correlated updates. In this paper, we sample transitions from $D$ uniformly at random as in \cite{dqn:scheme}. However, \cite{Schaul:scheme} and \cite{dudqn:scheme} show that having a prioritized sampling can lead to substantial improvement in performance.




\section{The Dynamic Frame skip Deep Q Network}

The key motivation behind the DFDQN architecture is the observation that when humans play games, the actions tend to be temporally correlated, almost always. For certain states of the game, we play long sequences of the same action whereas for some states, we switch the actions performed quickly. For example, in Seaquest, if the agent is low on oxygen and deep inside the sea, we would want the agent to resurface and fill up oxygen using a long sequence of {\it up} actions.
To incorporate the longer repetitive actions, we introduce the following architecture level changes to the DQN:\\
Let $\mathcal{A} = \{a_{1}, \cdots, a_{|\mathcal{A}|} \}$ denote the set of all legal actions of a particular Atari domain game (for Seaquest $\mathcal{A} = \{0,1,\cdots,17\}$). We introduce $|\mathcal{A}|$ new actions 
$\{ a_{|\mathcal{A}|+1}, \cdots, a_{2|\mathcal{A}|}\}$. The semantics associated with the actions are:
Action $a_{k}$ results in the same basis action being played in the ALE as the action  $a_{ (k\%|\mathcal{A}|)}$. The difference is in terms of the number of times the basis action is repeated by the ALE emulator. In \cite{dqn:scheme}, the DQN architecture operates with a static frame skip parameter which is equal to $4$. Hence any selected action is repeated $4$ times by ALE. In DFDQN, action $a_{k}$ is repeated $r_{1}$ number of times if $k <|\mathcal{A}|$ and $r_{2}$ number of times if $k \ge |\mathcal{A}|$.
One can think about each $a_{k}$ as representing a {\it temporally repeated} action. Given this, the objective of this work is to come up with a model to learn the near-optimal use of such options.
This scheme is implemented by doubling the number of neurons in the final layer of the DQN architecture. For a given state $s$, the DFDQN (with double the number of output neurons as DQN) thus outputs the discrete set of value function approximations $\{ Q(s, a_{1}), \cdots, Q(s, a_{2|\mathcal{A}|})\}$. $Q(s, a_{k})$ represents an approximation to the return the agent would get on taking action $a_{k}$ in state $s$ and following the optimal policy thereafter. Due to these additional repetitive actions, the value functions learnt by DFDQN differ drastically when compared to those learnt by the DQN (Figures 4 and 5).

\begin{figure}[H]
  \centering
  \includegraphics[width=.65\linewidth]{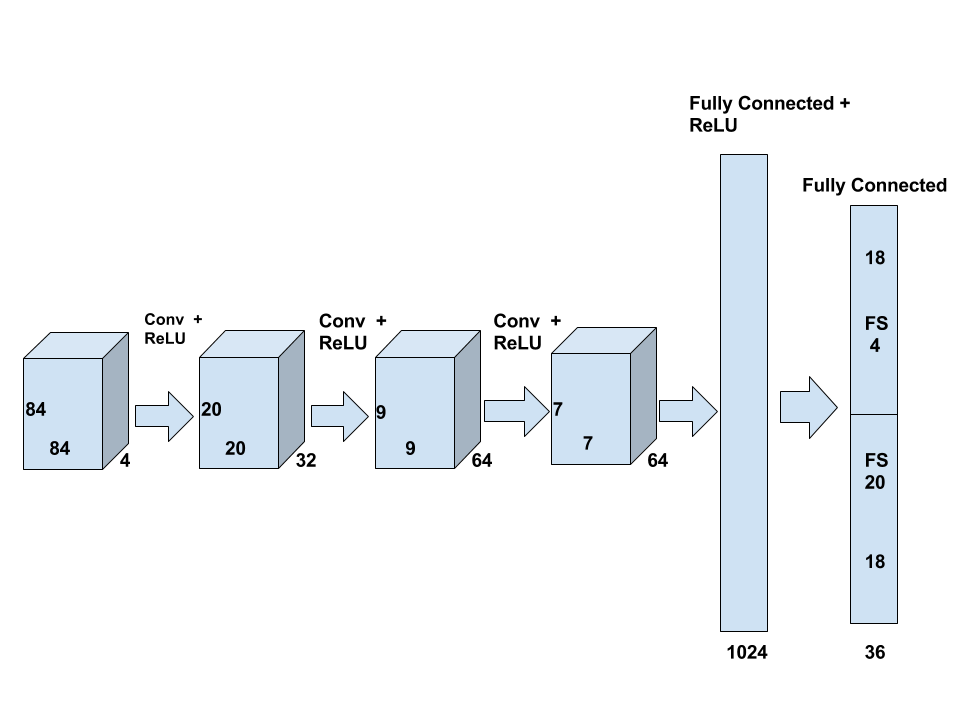}
  \caption{DFDQN Architecture}
  
\end{figure}

Note that $r_{1}$ and $r_{2}$ are hyper-parameters in this model and must be chosen before the learning begins. This architecture presents the agent with $2$ discrete levels of control, as far as action repetition is concerned. 

Ideally we would want to learn policies in a parametrized action space where the agent outputs an action $a_{k}$ and a parameter $r$ where $r$ denotes the number of times that the agent wants to play action $a_{k}$ consecutively. Such a framework would have the representative power to select the optimal number of times an action is to be executed. But, it would also be proportionately complex and as a result difficult to train. The DFDQN framework seeks to provide an approximation to this richer model and favors simplicity over optimality of game play. As future work, we wish to explore the learning of parametrized policies for games in the Atari $2600$ domain, learnt with the help of an actor-critic algorithm.  

\section{Experiments and Results}
 We perform general game-playing evaluation on three Atari $2600$ domain games, Seaquest, Space Invaders and Alien. We deploy a single algorithm and architecture, with a fixed set of hyper-parameters, to learn to play all three games given only raw pixel observations and game rewards. The ALE is a very challenging environment since some of the games such as Seaquest are reasonably complex and the observations used are high-dimensional (each frame is sampled as an image of dimensions $210 \times 160$).

 Our network architecture has the same low-level convolutional structure and input preprocessing of DQN \cite{dqn:scheme}. 
 Due to the partial observable nature of the Atari $2600$ games, the three previous frames are combined with the current frame and given as a $84\times84\times4$ multi-channel input to the DFDQN. This is followed by $3$ convolutional layers, which are in turn followed by $2$ fully-connected layers. The first convolutional layer has $32$ filters of size $8 \times 8$ with stride $4$; the second $64$ filters of size $4 \times 4$ with stride $2$, and the third convolutional layer consists of $64$ filters of size $3 \times 3$ with stride $1$. Since DFDQN has double the number of output neurons as DQN, we wanted more representational power for being able to decide from a larger set of actions. Therefore, we have $1024$ units in the pre-output hidden layer as compared to $512$ units used in the DQN architecture (\cite{dqn:scheme}). All the hidden layers ($3$ convolutional and $1$ fully connected) are followed by a rectifier nonlinearity.
 
\begin{figure}[H]
  \centering
  \includegraphics[width=.7\linewidth]{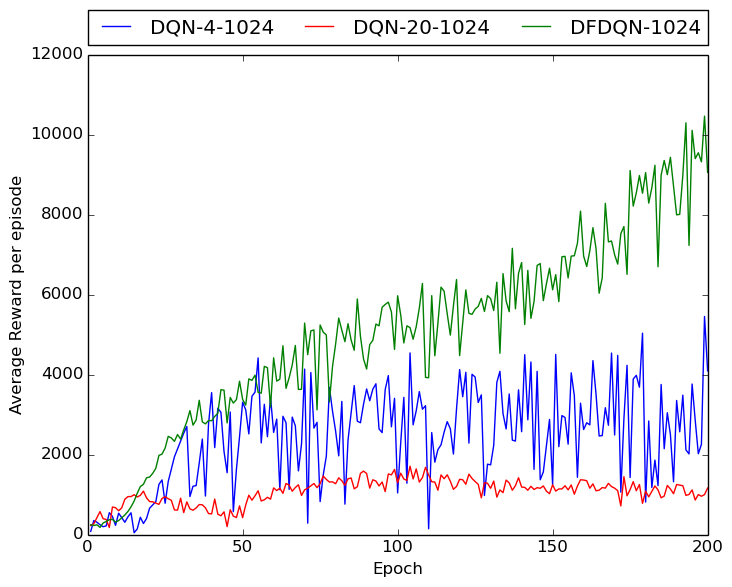}
  \caption{Plot of average reward per episode versus epoch number for the game of Seaquest. }
  \label{fig:sear}
  \end{figure}

\begin{figure}
  \centering
  \includegraphics[width=.7\linewidth]{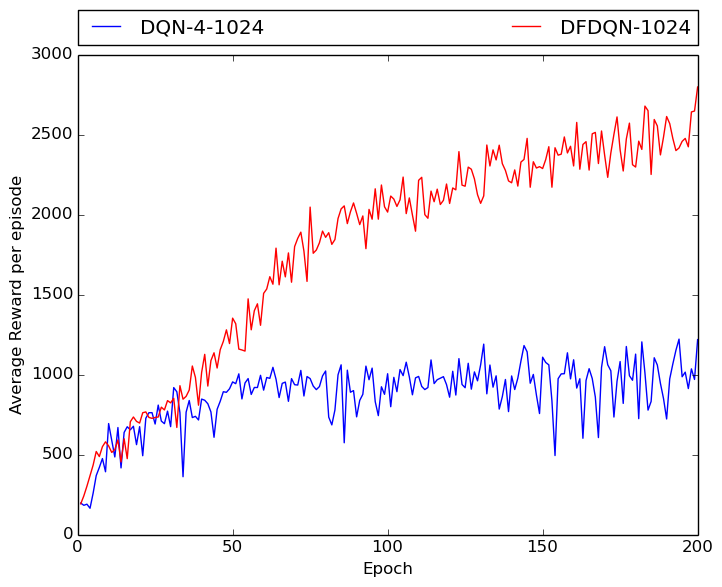}
  \caption{Plot of average reward per episode versus epoch number for the game of Space Invaders.}
\label{fig:spacr}
 \end{figure}

\begin{figure}
  \centering
 \includegraphics[width=.7\linewidth]{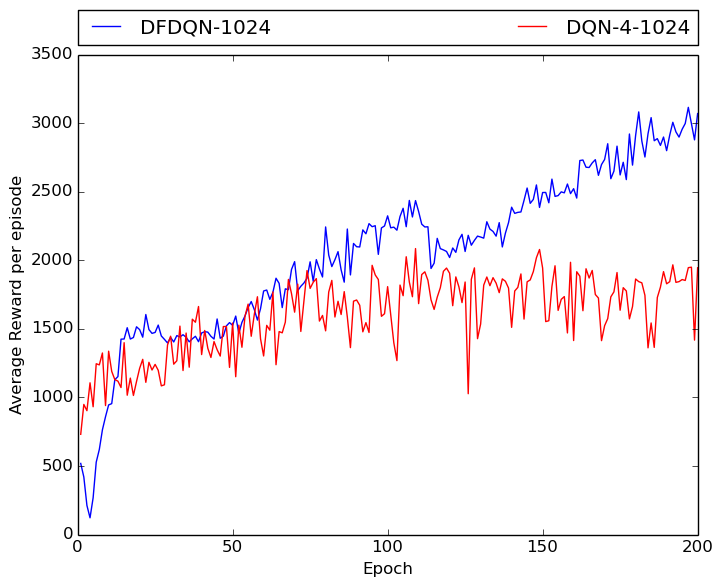}
  \caption{Plot of average reward per episode versus epoch number for the game of Alien.}
\label{fig:alienr}
  \end{figure}

\begin{figure}
  
   \centering
  \includegraphics[width=.7\linewidth]{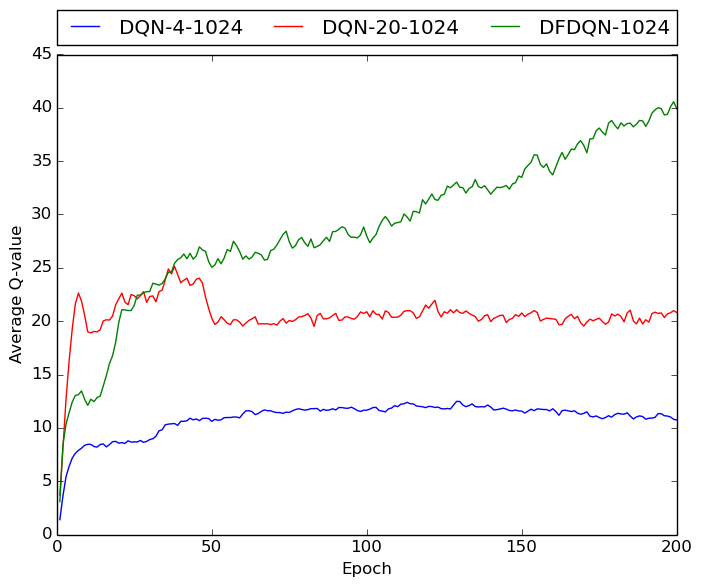}
  \caption{Plot of average Q-Value over all actions and steps versus epoch number for the game of Seaquest.}
\label{fig:seaq}
\end{figure}

\begin{figure}
  \centering
  \includegraphics[width=.7\linewidth]{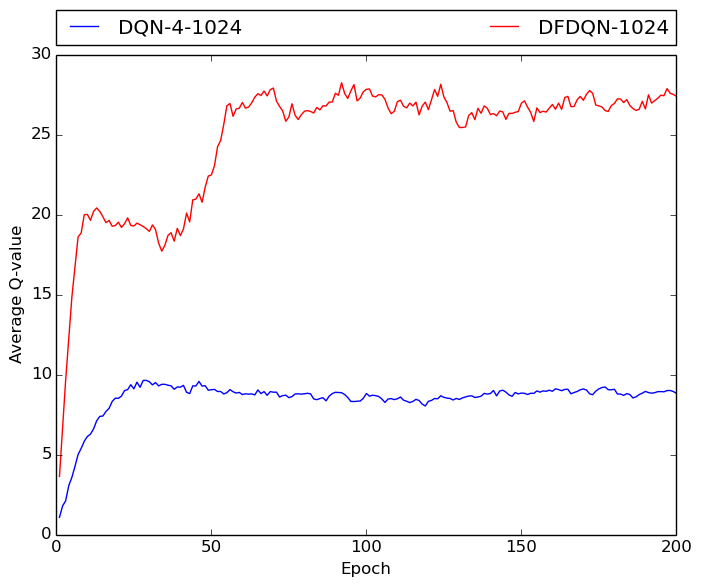}
  \caption{Plot of average Q-Value over all actions and steps versus epoch number for the game of Space Invaders.}
  \label{fig:spacq}
  \end{figure}

\begin{figure}
\centering
  \includegraphics[width=.7\linewidth]{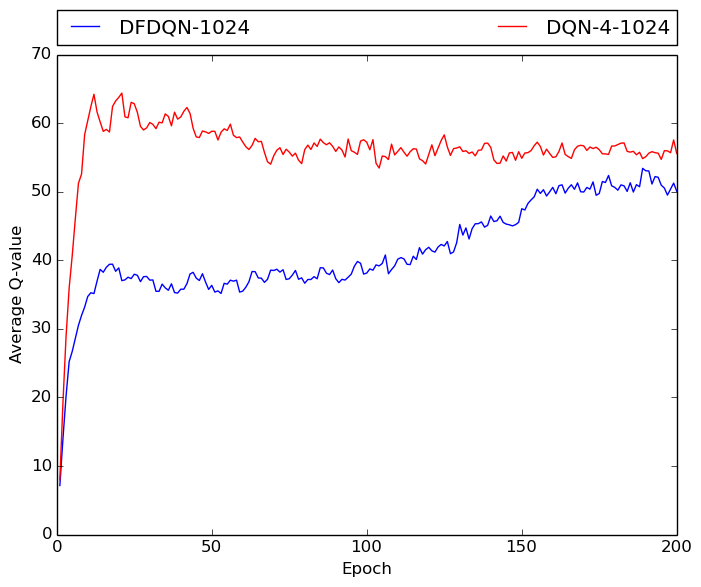}
  \caption{Plot of average Q-Value over all actions and steps versus epoch number for the game of Alien.}
\label{fig:alienq}
\end{figure}

 The values of $r_{1}, r_{2}$ (defined in the previous section) are kept the same for all three games and are equal to $4$ and $20$ respectively. Similar to \cite{dqn:scheme}, the hyper-parameters of the DFDQN are kept the same for all three games. We have double the number of actions as compared to DQN. Thus, to ensure sufficient exploration, we anneal the exploration probability $\epsilon$ from $1$ to $0.1$ over $2$ million steps as against $1$ million steps used in DQN. 
 We again follow \cite{dqn:scheme} in using RMSProp with a max-norm clipping of $1$ for our gradient updates while learning.
 Even though our Q values are larger due to extended options possibly giving higher rewards, we still use a max-norm clipping of $1$ as used in DQN. We leave it for future work to find the best max-norm clip value, as well as other hyper-parameters like learning rate, final exploration probability $\epsilon$, replay memory size, freeze interval, etc that would work well for all games in the DFDQN setup.
 To establish the fact that the improvement in performance is not just due to the increase in the representational power by having double the number of pre-final hidden layer neurons, we run DQN baselines with $1024$ units in the pre-final layer for all three games. The DQN implementation we use is: \url{https://github.com/spragunr/deep_q_rl} and is based on Theano \cite{Bastien-Theano-2012,bergstra+al:2010-scipy}.
 We also report the scores obtained from original DQN architecture with $512$ pre-final hidden layer neurons from the most recent usage of DQN in \cite{dudqn:scheme}, where DQN is reported as baseline for their DuDQN model (the current state-of-the-art model for Atari $2600$ domain).
 The training and evaluation framework is the same as that used in \cite{dqn:scheme}. A training epoch consists of $250000$ steps (action selections). This is followed by a testing epoch which consists of $125000$ steps. The score of a single episode is defined to be the sum of the rewards received from the ALE emulator in that episode. The score of a testing epoch is defined to be the average of the scores obtained in all of the complete episodes played during that testing epoch. The scores reported here are for the testing epoch with the highest average episode score (known as the {\it best score} for a game). 
 Table $1$ presents the experimental results. HLS denotes the pre-final hidden layer size. FS stands for the frame-skip value used. A value of $D$ for FS stands for dynamic. Arch denotes the architecture used. AS denotes the best average testing epoch score as defined above. \\
 In each of the following figures, one epoch consists of $125000$ steps (decisions). DQN-$a$-$b$ refers to DQN architecture that has $b$ units in the pre-output layer and operates with frame skip rate of $a$. DFDQN-$b$ is the Dynamic Frame skip Deep Q-Network architecture with $b$ units in pre-final layer. \\
 Figure \ref{fig:sear}, Figure \ref{fig:spacr} and Figure \ref{fig:alienr} depict the change in the expected cumulative reward per testing episode with time for the games of Seaquest, Space Invaders and Alien respectively. DFDQN with $1024$ units in the pre-output layer significantly outperforms DQN with $1024$ neurons in pre-output layer. 
 Figure \ref{fig:seaq}, Figure \ref{fig:spacq} and Figure \ref{fig:alienq} depict the evolution of the average Q-values over all the actions with time for the games of Seaquest, Space Invaders and Alien respectively. 
 \begin{table}[H]
\centering
\begin{tabular}{|l|l|l|l|l|l|}
\hline
               
S.No.  & Game & HLS & FS & Arch & AS    \\ \hline
1  & Seaquest  & 1024 & 4  & DQN  & 5450 \\ \hline
2  & Seaquest  & 1024 & 20 & DQN  & 1707 \\ \hline
3  & Seaquest  & 1024 & D  & DFDQN  & \bf{10458} \\ \hline
4  & Seaquest  & 512  & 4  & DQN  & 5860 \\ \hline
5  & Space Invaders& 1024 & 4 & DQN  & 1189 \\ \hline
6  & Space Invaders  & 1024 & D & DFDQN  & {\bf 2796} \\ \hline
7  & Space Invaders  & 512 & 4 & DQN  & 1692 \\ \hline
8  & Alien  & 1024 & 4 & DQN  & 2085 \\  \hline
9  & Alien  & 1024 & D & DFDQN  & {\bf 3114} \\ \hline 
10 & Alien  & 512 & 4 & DQN  & 1620 \\  
\hline
\end{tabular}
\caption{Experimental results}
\end{table}


 An interesting characteristic of the graph (Figure 4) is that the DFDQN-$1024$'s Q-value for Seaquest keeps increasing with time even at the end of $200$ epochs. This means that there is still scope for the performance to improve on further training beyond $200$ epochs. To verify our claim that temporally extended actions can lead to better policies, we did an analysis of the percentage of times the longer sequence of basis actions was selected.
 Using the network which yielded the {\it best score} (notion of best score defined in the previous section), we ran a testing epoch of $10000$ decision-making steps, which resulted in $17$ episodes for Seaquest, $18$ episodes for Space Invaders and $18$ episodes for Alien. We recorded the actions executed at each decision-step. On an average (over the completed episodes), the DFDQN chooses the longer (frame skip $20$) actions $69.99\%$ of the times in Seaquest, $78.87\%$ of the times in Space Invaders and $71.31\%$ of the times in Alien. This shows that the DFDQN agent is able to make good use of the longer actions most of the times, but does not ignore the shorter actions. The agent has learnt through repeated exploration-feedback cycles to prefer the extended actions but still exercises the fast reflexes when the situation needs it to do so. We can safely claim that the addition of these higher frame skip options has been an important contributing factor to the improvement in performance. To strengthen our claim that there is a need for {\it dynamic} frame skip, we show results on one game (Seaquest) that DQN with a purely high but {\it static} frame skip of $20$ scores poorly on gameplay.
 Videos of some of the learned policies for Space Invaders, Seaquest and Alien are available at 
\url{https://goo.gl/VTsen0}, \url{https://goo.gl/D7twZc} and \url{https://goo.gl/aCcLb7 }) respectively.


\section{Conclusions}

In this paper, we introduce a new architecture called Dynamic Frame skip Deep Q-Network (DFDQN). It introduces temporal abstractions through extended actions without losing the ability to make quicker reflexes when required. We show empirically that DFDQN leads to significant improvement in the performance compared to DQN with results on three Atari $2600$ domain games over DQN : Seaquest, Space Invaders and Alien. The results section illustrates the importance of the extended actions with the temporally extended versions of the actions being selected $69\%$ of the times in Seaquest, $78\%$ of the times in Space Invaders and $71\%$ of the times in Alien.

\section{Discussion}

Our experiments were conducted with frame skip rates of $4$ and $20$. A generic framework would enable the agent to play with a wide range of frame skip rates selected according to the current state of the game. 
We outline such a generic framework with a structured policy wherein the agent does not just decide the action but also how long the action should persist. For this, we would need an Actor Critic setup similar to \cite{Hausknecht:scheme}. \cite{Hausknecht:scheme} use Deep Actor Critic for learning parametrized policies in the Half Field Offense (Robo Soccer) domain. Their policy is composed of continuous valued parameters like {\it Power, Direction} in addition to the action probabilities of {\it Kick, Turn, Tackle, Dash}. The parameters are restricted to lie in a fixed continuous interval. We can adapt this setup for learning structures policies in the Atari $2600$ domain with the continuous valued parameter being the frame skip rate $r$.

In this setup, we envision the network as being composed of three parts: The core-policy sub-network $N_{c}$, a parameter sub-network $N_{p}$ and the critic sub-network $N_{cr}$. All three of them share the same lower level convolutional filters up to a desired depth similar to the architecture used in \cite{a3c:scheme} which as far as we know, is the only successful attempt at using Actor Critic algorithms in the Atari $2600$ domain. 
The output of $N_{c}$ is the set of action probabilities  $\{ \pi_{1}, \pi_{2}, \cdots, \pi_{|\mathcal{A}|} \}$ where $\pi_{k}$ is the probability of choosing action $a_{k}$ and $\mathcal{A} = \{ a_{1}, a_{2}, \cdots, a_{|\mathcal{A}|}\}$ is the set of basis actions for the chosen task. $N_{p}$ has a single scalar output which is the frame skip parameter $r$. We could constrain $r$ to lie in a fixed interval $[r_{\textrm{min}},r_{\textrm{max}}]$, where $r_{\textrm{min}}$ could be $1$ and $r_{\textrm{max}}$ could be as high as $100$. $N_{c}$ and $N_{p}$ together constitute the Actor part of the Actor Critic network. The Critic part depicted by $N_{cr}$ outputs a single scalar value $v$ which is an estimate of the Value Function of the current state. The error function to be optimized and the manner in which gradients are to be backpropagated can closely follow the Actor-Critic implementation given in \cite{a3c:scheme}. 

\begin{figure}[H]
  \centering
  \includegraphics[width=.8\linewidth]{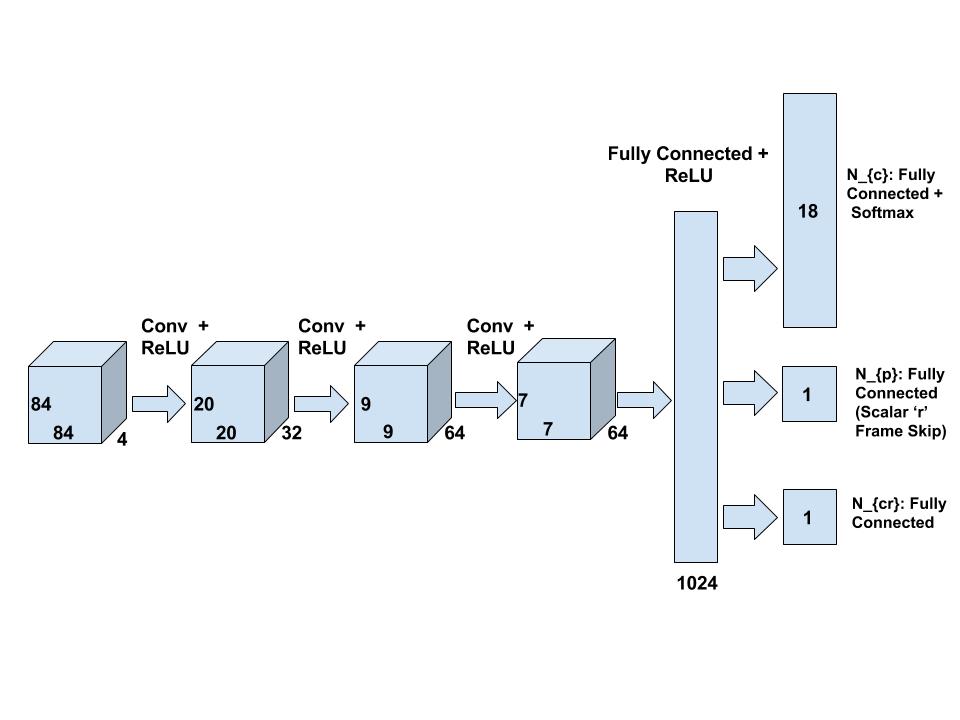}
  \caption{Proposed Architecture}
\label{fig:arch2}
\end{figure}

An alternative way of learning to perform repeated actions could be to use the previous action as input to the policy similar to \cite{riddle:scheme}. This could be combined with the proposed Actor Critic setup to help on deciding the frame skip parameter $r$.\\
The improvement in performance has been achieved without much tuning of the DQN (\cite{dqn:scheme}) network parameters. We believe the DFDQN setup will further improve with more appropriate hyper-parameter choices. As future work, we would also like to explore the DF paradigm on more games in the Atari $2600$ domain with the more recent state of the art methods such as Prioritized Replay \cite{Schaul:scheme}, Dueling DQN \cite{dudqn:scheme} and Asynchronous-DQN \cite{a3c:scheme}. 
We believe that a combination of Dynamic Frame skip and Duelling DQN (DFDuDQN) and (or) Prioritized Replay would lead to a state of the art model for games in the Atari $2600$ domain.


\section*{Acknowledgements:}

We thank the reviewers of the IJCAI-16 Workshop on Deep Reinforcement Learning: Frontiers and Challenges for their useful comments regarding our work. We also thank Karthik R. Narasimhan (CSAIL MIT) and Leandro Soriano Marcolino (Teamcore USC) for their suggestions and discussions.




\bibliographystyle{named}
\bibliography{ijcai16}

\end{document}